\pdfoutput=1
\PassOptionsToPackage{table}{xcolor}
\documentclass[11pt]{article}

\usepackage{ACL2023}

\usepackage{times}
\usepackage{latexsym}
\usepackage{hyperref}
\usepackage{booktabs}
\usepackage{colortbl}
\usepackage{xcolor}
\usepackage{tabularx}
\usepackage{graphicx}
\usepackage{makecell}
\usepackage{multirow}
\usepackage{multicol}
\usepackage{rotating}
\usepackage{balance}
\usepackage{tablefootnote}

\usepackage{threeparttable}

\usepackage[T1]{fontenc}

\usepackage[utf8]{inputenc}
\usepackage{csquotes}

\usepackage{microtype}

\usepackage{inconsolata}

\newcommand{\minus}{\scalebox{0.6}[1.0]{$-$}}

\title{Language Models for German Text Simplification:\\ Overcoming Parallel Data Scarcity through Style-specific Pre-training}

 \author{Miriam Ansch\"{u}tz, Joshua Oehms, Thomas Wimmer, \\
{\bf Bart\l{}omiej Jezierski} \and {\bf Georg Groh} \\
         School for Computation, Information and Technology \\
         Technical University of Munich, Germany\\
         \texttt{\{miriam.anschuetz, joshua.oehms, thomas.m.wimmer, b.jezierski\}@tum.de}\\ \texttt{grohg@in.tum.de}
         }

\begin{document}
\maketitle
\begin{abstract}
Automatic text simplification systems help to reduce textual information barriers on the internet. However, for languages other than English, only few parallel data to train these systems exists. We propose a two-step approach to overcome this data scarcity issue. First, we fine-tuned language models on a corpus of German Easy Language, a specific style of German. Then, we used these models as decoders in a sequence-to-sequence simplification task. We show that the language models adapt to the style characteristics of Easy Language and output more accessible texts. Moreover, with the style-specific pre-training, we reduced the number of trainable parameters in text simplification models. Hence, less parallel data is sufficient for training. Our results indicate that pre-training on unaligned data can reduce the required parallel data while improving the performance on downstream tasks.
\end{abstract}

\section{Introduction}
Automatic text simplification (ATS) is the task of simplifying a text's lexical and structural complexity while preserving its original meaning. Easy-to-read texts can help people with learning deficiencies or non-native speakers gain access to texts that they could not understand otherwise. On the one hand, ATS can be used to create assisting tools for people with reading disabilities or professional translators \citep{Suarez-ATS-for-E2R-experts}.
On the other hand, ATS can be applied as a preprocessing step for other natural language processing tasks such as machine translation or information retrieval to improve their performances \citep{Stajner-ATS-for-MT}, making it an important field of study.

In German, there exist multiple levels of simplified language. In contrast to the underspecified simple language, the so-called Leichte Sprache (Easy Language) enforces a very strong simplification level and follows predefined structural rules \citep{Leichte-Sprache-Regeln}. These rules include conveying only one message per sentence (structural simplification), restriction to common words (lexical simplification), and usage of simplified grammar (syntactical simplification). This simplified grammar breaks with standard German grammar, for example, by using dative instead of genitive to indicate possession. We consider Easy Language as a standalone language style. Therefore, we refer to Easy Language data as monolingual data in the further course of the paper, even though it is German as well.

This work shows the benefits of fine-tuning language models for specific styles and characteristics. We publish and discuss a collection of causal language models fine-tuned for German Easy Language. As shown in previous work \citep{Gururangan-pretraining-benefits}, pre-training language models for specific domains can benefit the performances of downstream tasks in the respective domain. We extend this analysis to the language style of Easy Language. In addition, the fine-tuned models can be used to generate text with the specificities of Easy Language, for example, in data augmentation applications. Finally, we present how these models can serve as plug-in-decoders in BART-like architectures \citep{Lewis-BART} to speed up and improve the training on sequence-to-sequence (seq2seq) tasks. Therefore, our contributions are the following:
\begin{itemize}
    \item We publish five German Easy Language causal language models and extensively evaluate their language style adaptions.
    \item We assess the models' performance on the two downstream tasks of text complexity prediction and text simplification.
    \item We suggest an ATS training process that exploits our pre-trained language models. This process reduces the number of trained parameters by over 90\% while preserving state-of-the-art performance.
\end{itemize}
With the reduction of trainable parameters, less aligned data is needed to train an ATS system. Especially for languages other than English, where aligned data is sparse, pre-trained causal language models can improve ATS performance.
We publish our code and results for further research and application\footnote{
\url{https://github.com/MiriUll/Language-Models-German-Simplification}
}.


\section{Related work}\label{sec:rel_work}
Causal language models can complete text based on a prompt. In contrast to masked language models, where the models know about the context before and after a specific token, these causal language models rely only on the input and the previously outputted tokens. Therefore, they are called autoregressive models. The Generative Pre-trained Transformer (GPT) \citep{Radford-GPT2} is a prominent example of such an autoregressive language model. It was trained on a collection of web data and, thus, outputs text for general purposes. Previous work has fine-tuned GPT for multiple domains and tasks, such as the task of quest generation in games \citep{Värtinen-Quest-GPT} or the medical domain \citep{Schneider-Portuguese-medical}. In addition to domain adaption, GPT was tailored to specific text styles and characteristics. These style transfer approaches include fine-tuning for poem generation \citep{Liao-gpt-poem} or the reduction of non-normative clauses \citep{Peng-non-normative}. \citet{Li-GPT-D} trained a GPT model to mimic the language of people with dementia. By calculating the perplexities of texts with the fine-tuned and original version, they could distinguish samples from healthy and diseased people.

\citet{Sun-SimpleBERT} adapted a language model for simple language by only masking easy-to-understand words in training. However, this model is a masked language model that can only fill in blanks and not generate text from scratch. Most similar to our work is the TransformerLM by \citet{Maruyama-ats-transformerLM} trained for Japanese text simplification. The authors used a parallel corpus to directly fine-tune a GPT model for simplification. In contrast, our models are fine-tuned on monolingual Easy Language data. Therefore, they do not require alignments and can be used for a broader range of tasks.

\subsection{German Text simplification}
In contrast to the English language, automatic text simplification in German has seen little research. The first system for Easy Language was proposed by \citet{Suter-rule-based-German} and consisted of a collection of hand-crafted rules, including sentence splitting and paraphrasing. \citet{Sauberli-first-NMT} published the first neural simplification approach based on the transformer architecture, together with an aligned corpus. They discussed multiple data augmentation strategies, but their results lacked fluency and content preservation. Based on an extended version of this dataset, \citet{Spring-multi-level-German} built a controllable simplification system that can output different simplification levels based on the Common European Framework of References for Languages (CEFR), but not specifically Easy Language. Finally, \citet{rios2021newdata} proposed a modified mBART architecture for document-level simplification. In our paper, we adopted their architecture to evaluate our language models on the downstream task of ATS.

\section{Datasets}\label{sec:datasets}
Several sources are available in Easy Language; however, they mostly encompass news websites, and only a few are aligned with articles in standard German. In the following sections, we detail the information on the data used in our training, including the Easy Language monolingual corpus utilized for fine-tuning German language models and the parallel corpus for the downstream task of text simplification. The dataset utilized for the downstream task of text complexity prediction is publicly available as a part of the GermEval 2022 shared task \citep{Mohtaj-GermEval22} (refer to Subsection \ref{subsec:text-complexity}). 
We published scrapers to recreate our sources for the use of the academic community\footnote{\url{https://github.com/brjezierski/scrapers}}. We also provide an overview of available monolingual and parallel data sources for simplified German beyond our training data in Appendix~\ref{sec:appendix}.

\subsection{Monolingual corpus}
\label{sec:monolingual_corpus}

An overview of the available monolingual data can be found in Table \ref{tab:monolingualDataset}. The publicly available Easy Language datasets are very limited: The Simple German corpus published by \citet{Toborek2022Sep} contains texts on health and medication, public administration, politics, information texts for disabled people, and news articles.
The second publicly available resource is a small corpus published by \citet{trondheim:2018}. 
It contains election programs, excerpts from the Bible, children's stories, and Red Cross documents.

Kurier, InfoEasy, and NDR are public broadcasting services in Austria, Switzerland, and northern Germany, respectively, and have specific columns in Easy Language. In addition, Hurraki and Lebenshilfe offer online dictionaries in Easy Language, while Einfachstars contains news articles about celebrities. These three data sources diversify our covered domains and styles of writing. More details about the data sources can be found in Table \ref{tab:monolingualDataOverview} in Appendix \ref{sec:appendix}. Our fine-tuning data combines all sources included in Table \ref{tab:monolingualDataset}. The combined data was shuffled and randomly split into a training set containing 90\% of the data and a validation set with 10\% of the total. 

\begin{table}[ht]
\centering
\rowcolors{2}{gray!10}{white}
\begin{tabular}{lrl}
\toprule
\textbf{Dataset} & \textbf{Sentences} & \textbf{Domain}\\
\midrule
 Hurraki              & 56,785      & lexicon \\
 Lebenshilfe          & 7,144       & lexicon   \\
 Einfachstars         & 129,674     & news    \\
 Nachrichtenleicht    & 122,842     & news    \\
 Kurier               & 67,827      & news    \\
 NDR                  & 60,749      & news    \\
 InfoEasy             & 10,310      & news     \\
 \citet{trondheim:2018} & 4,210       & misc.   \\ 
 \citet{Toborek2022Sep} & 28,356      & misc.     \\
 \midrule
 \textbf{Total}              & \textbf{544,467}     &  \\
 \bottomrule
\end{tabular}
\caption{Overview of the monolingual data used for language model fine-tuning.}
\label{tab:monolingualDataset}
\end{table}

\subsection{Parallel corpus}
For training the text simplification model, we used the publicly available 20 Minuten dataset\footnote{\url{https://github.com/ZurichNLP/20Minuten}}. The dataset consists of full articles paired with shortened, simplified summaries from the Swiss news magazine 20 Minuten. It comprises 17,905 article pairs in the training dataset and 200 pairs in the validation and test set each \citep{rios2021newdata}. The dataset's compression ratio (the reduction in the word count of simplified summaries) was estimated at 11\%. 

\subsection{Preprocessing pipeline}
\label{sec:preprocessing}
Analyzing the outputs of publicly available language models in standard German, we noticed that in many cases, especially for the news headline-like input, the output contained noise, such as HTML tags or URLs. For this reason, coupled with the fact that we obtained data from multiple sources using various formats, we built a shared preprocessing pipeline to standardize the input for the fine-tuning of the language models as well as the simplified parts in the aligned dataset. Our pipeline removed redundant tags and characters. Some Easy Language texts use bullet points to break down sentences. Since most of the data did not follow this guideline, we converted the existing bullet points into comma-separated phrases. Another feature of Easy Language is the hyphenation of compound nouns. We compiled a list of hyphenated nouns in the monolingual dataset and used it to replace equivalent non-hyphenated compound nouns. 

\section{Methodology}\label{sec:methodology}
Our approach is divided into two parts. First, we fine-tuned generative language models for German Easy Language. Then, we used these models as plug-in decoders in a BART-based simplification task.
\subsection{Fine-tuning language models}
\begin{table*}[ht]
    \centering
    \rowcolors{2}{gray!10}{white}
    \begin{tabularx}{\linewidth}{lXlc}\toprule
        \textbf{Model} & \textbf{Training data} & \textbf{Initialization} & \textbf{\#Params}\\\midrule
        GerPT2 \citep{Minixhofer-gerpt2} & CC-100 Corpus & English GPT2 &163M\\
        german-gpt2 \citep{Schweter-german-gpt2} & Wikipedia dump, EU Bookshop corpus, Open Subtitles, CommonCrawl, ParaCrawl and News Crawl & from scratch &124M\\
        \makecell[Xt]{GPT2 Wechsel\\\hspace{2mm} \citep{Minixhofer-wechsel}} & OSCAR corpus, MUSE &English GPT2 &124M\\
        Oscar fine-tune \citep{gpt2-oscar} & OSCAR corpus & \textit{no info} &354M\\
        \makecell[Xt]{mGPT \citep{Shliazhko-mGPT}\\\hspace{2mm} \textit{(multilingual)}} & Wikipedia, Colossal Clean Crawled Corpus& from scratch &1417M\\
        \bottomrule
    \end{tabularx}
    \caption{Training setup and number of parameters for different German GPT2 models. These models were used as base for our Easy Language fine-tuning.}
    \label{tab:orig_infos}
\end{table*}
We selected five different pre-trained GPT-based models from Huggingface \citep{Wolf-Huggingface} as the base for our language models, four German models, and one multilingual model. As shown in Table \ref{tab:orig_infos}, the models differ in their original training data, initialization, and size. All German models use an embedding size of 1024, while mGPT has a size of 2048. To fine-tune the models, we used 
a NVIDIA A100 GPU. We trained for one epoch, with a learning rate of $1e^{\minus4}$, a weight decay of $0.01$, and a batch size of eight together with a gradient accumulation of four. However, due to the large model size, we had to decrease the batch size to one for mGPT. The dropout parameters for the embedding, the attention mechanism, and the fully connected layers were set to $0.1$ each. 

\citet{Su-contrastive-loss} proposed a new learning objective for generative language models, the contrastive loss. This loss adds a similarity regularization to the cross entropy loss to enforce discriminative token representations. We used this loss function together with an AdamW optimizer for our fine-tuning.

\subsection{Text simplification}
The simplification task can be considered as a translation-like seq2seq problem. Thus, we used an encoder-decoder architecture based on mBART's architecture \citep{Liu-mBART}. It consists of a BERT-like encoder and a GPT-like decoder. Additionally, mBART was pre-trained on multilingual data (including German) on a denoising objective 
and forms the current baseline for transformer-based German ATS \citep{rios2021newdata}. The baseline's mBART-encoder was modified to use sliding attention to be applied to article inputs. Thus, it was possible to use long input sequences efficiently. 
We adapted this architecture and replaced the mBART-decoder with our fine-tuned GPT models. For the target text, we used the same preprocessing used for fine-tuning the decoder models. As our language models already output text in the desired style, no further training of the decoder was necessary. Therefore, we only trained the encoder-decoder cross attention to align the encoding of the complex articles with our language models. This was proven successful for machine translation with pre-trained language models by \citet{gheini2021cross}. Training only the cross attention reduced the number of parameters to be updated, making the training of the simplification more efficient. In addition, the language models were not updated, and thus, we avoided catastrophic forgetting \citep{Goodfellow-catastrophic-forgetting} of their German language comprehension. We trained with the same hyperparameters as the baseline, except we set label smoothing to zero and added a contrastive part to the loss function \citep{Su-contrastive-loss}.
We trained on a single NVIDIA TITAN X. Similar to the baseline, the training converged after 3 to 4 days according to validation loss, which means training for about 20 epochs.
Due to hardware limitations, we trained with a batch size of one and a gradient accumulation of 32.

\section{Evaluation}\label{sec:eval}
This section describes four experiments to compare our fine-tuned (FT) models with their original (O) versions. First, we measured the models' perplexities on easy and normal texts and analyzed the readability of their outputs. In addition, the models were evaluated on two downstream tasks; text complexity prediction and automatic text simplification.
\subsection{Perplexity scores}
The perplexity describes how likely a specific model will produce a given text. A lower perplexity score indicates a better match between the model and text. 
We evaluated how well our models adapt to the style of Easy Language. Therefore, the fine-tuned and original models' perplexities on easy and normal texts were compared. The data was collected from the MDR, a public broadcasting service in Germany that publishes news articles in Easy Language. We manually aligned 100 paragraphs from the easy and original articles. To calculate the perplexity of the data, we used the tutorial code from Huggingface \citep{Huggingface-perplexity} that implements perplexity as a sliding window over the input data. We adapted the code for a sample-wise calculation and averaged the perplexity over all samples.

Perplexity is highly dependent on the tokenization and the length of the samples \citep{Wang-PPL-critique}. Therefore, we cannot determine the best fine-tuned models by selecting the model with the lowest perplexity. However, the fine-tuned and original versions of the models use the same tokenizers. Thus, we can compare their perplexities and assess the effects of fine-tuning.

Table \ref{tab:ppl_comparison} shows the average perplexity values for the easy and normal texts. No model has seen any of the data before in training. All fine-tuned models show a lower perplexity for the Easy Language samples. In contrast, except for one model, the original models perform better on the normal texts. This suggests that the fine-tuned models match the specificities and structure of Easy Language better and, thus, that they are more likely to produce similar texts.\\
\begin{table}[ht]
    \centering
    \rowcolors{2}{gray!10}{white}
    \begin{tabular}{lcc|cc}
    \toprule
        & \multicolumn{2}{c|}{\textbf{Easy text}} & \multicolumn{2}{c}{\textbf{Normal text}}\\
        \multirow{-2}{*}{\textbf{Model}} &\textbf{FT}&\textbf{O}&\textbf{FT}&\textbf{O}\\\midrule
        gerpt2 & \textbf{25.35}& 51.31& \textbf{53.74}& 56.42\\
        german\_gpt & \textbf{31.81}& 47.19& 77.76& \textbf{31.49}\\
        wechsel & \textbf{25.99}& 38.98& 69.29& \textbf{34.80}\\
        oscar & \textbf{34.24}& 59.31& 112.75& \textbf{66.22}\\
        mGPT & \textbf{24.93}& 25.05& 99.53& \textbf{19.18}\\
    \bottomrule
    \end{tabular}
    \caption{Comparison of perplexity scores between easy and normal texts. Lower score means better match. The fine-tuned models fit easy German text better, while the original models favor normal texts.}
    \label{tab:ppl_comparison}
\end{table}

\subsection{Readability and Easy Language characteristics}\label{sec:promots}
To evaluate the readability of the models' outputs, we compared the Flesch Reading Ease (FRE) scores \citep{Amstad-FRE} of sample outputs. We prompted the models with six different inputs: \enquote{Das}(\textit{This}), \enquote{Heute}(\textit{Today}), \enquote{Wir}(\textit{We}), \enquote{Die T{\"u}rkei}(\textit{Turkey}), \enquote{Dieses Haus}(\textit{This house}), and \enquote{Mein Vater}(\textit{My father}). The models had to output 100 new tokens, and we set a repetition penalty to enforce novel content in the output. Moreover, three different decoding strategies (contrastive search, sampling, and beam search) were used, resulting in 18 output texts per model. Finally, the FRE score was calculated for each of the model outputs. This score considers the average sentence length and the average number of syllables per word, which favors concise sentences with short words. Therefore, a higher score indicates a more accessible text. Table \ref{tab:readability_comparison} shows each model's average FRE score. The fine-tuned models achieve a higher score, which implies that their output is more readable than their original's. In addition, we counted the number of suggested newline (\textbackslash n) tokens. As presented in Table \ref{tab:readability_comparison}, the fine-tuned models output this token more often. This shows that they adapted to the Easy Language characteristic of only writing one thought per line.
\begin{table}[ht]
    \centering
    \rowcolors{2}{gray!10}{white}
    \begin{tabular}{lcc|cc}\toprule
         & \multicolumn{2}{c|}{\textbf{Average FRE}} & \multicolumn{2}{c}{\textbf{\textbackslash n tokens}}\\
        \multirow{-2}{*}{\textbf{Model}} & \textbf{FT} & \textbf{O} & \textbf{FT} & \textbf{O}\\ \midrule
        gerpt2 & \textbf{65.17} & 51.09 & \textbf{67} & 34\\ 
        german\_gpt & \textbf{75.09} & 70.89 & \textbf{79} & 74 \\ 
        wechsel & \textbf{70.72} & 55.86 & \textbf{69} & 18\\ 
        oscar & \textbf{68.21} & 49.32 & \textbf{61} & 0\\ 
        mGPT & \textbf{72.16} & 55.30 & \textbf{106} & 29\\ 
        \bottomrule
    \end{tabular}
    \caption{Flesch Reading Ease score averaged over different prompts and decoding strategies, and total number of \textbackslash n tokens suggested. The fine-tuned models output more simple texts.}
    \label{tab:readability_comparison}
\end{table}

To further investigate this conformity with Easy Language, we gave the models the input sentence \enquote{Heute scheint die Sonne} (\textit{Today sun is shining}) and let them predict the next token. As highlighted in Table \ref{tab:sentence_continuation}, most of the fine-tuned models proposed to end the sentence, i.e., predicted a point or a modifier. In contrast, the original models added further information by continuing the sentence with a comma or an \enquote{and}. 
\begin{table}[ht]
    \centering
    \rowcolors{2}{gray!10}{white}
    \begin{tabular}{lccc}\toprule
         & \multicolumn{3}{c}{\textbf{Suggested next token}} \\
        \multirow{-2}{*}{\textbf{Model}} && \textbf{FT} & \textbf{O}\\ \midrule
        gerpt2 &&\textbf{ .} & \textbf{,}\\
        german\_gpt && sehr (\textit{very}) &\textbf{,}\\
        wechsel && \textbf{.} & und (\textit{and})\\
        oscar && \textbf{.} & \textbf{,}\\
        mGPT && auf (\textit{on}) & bei (\textit{at})\\
        \bottomrule
    \end{tabular}
    \caption{Suggested next token for the input sentence \enquote{Heute scheint die Sonne} (\textit{Today the sun is shining}). The original models propose to continue the sentence, while the fine-tuned models only put one thought per sentence.}
    \label{tab:sentence_continuation}
\end{table}
\subsection{Human grammar evaluation}
\begin{figure}
    \centering
    \includegraphics[width=\linewidth]{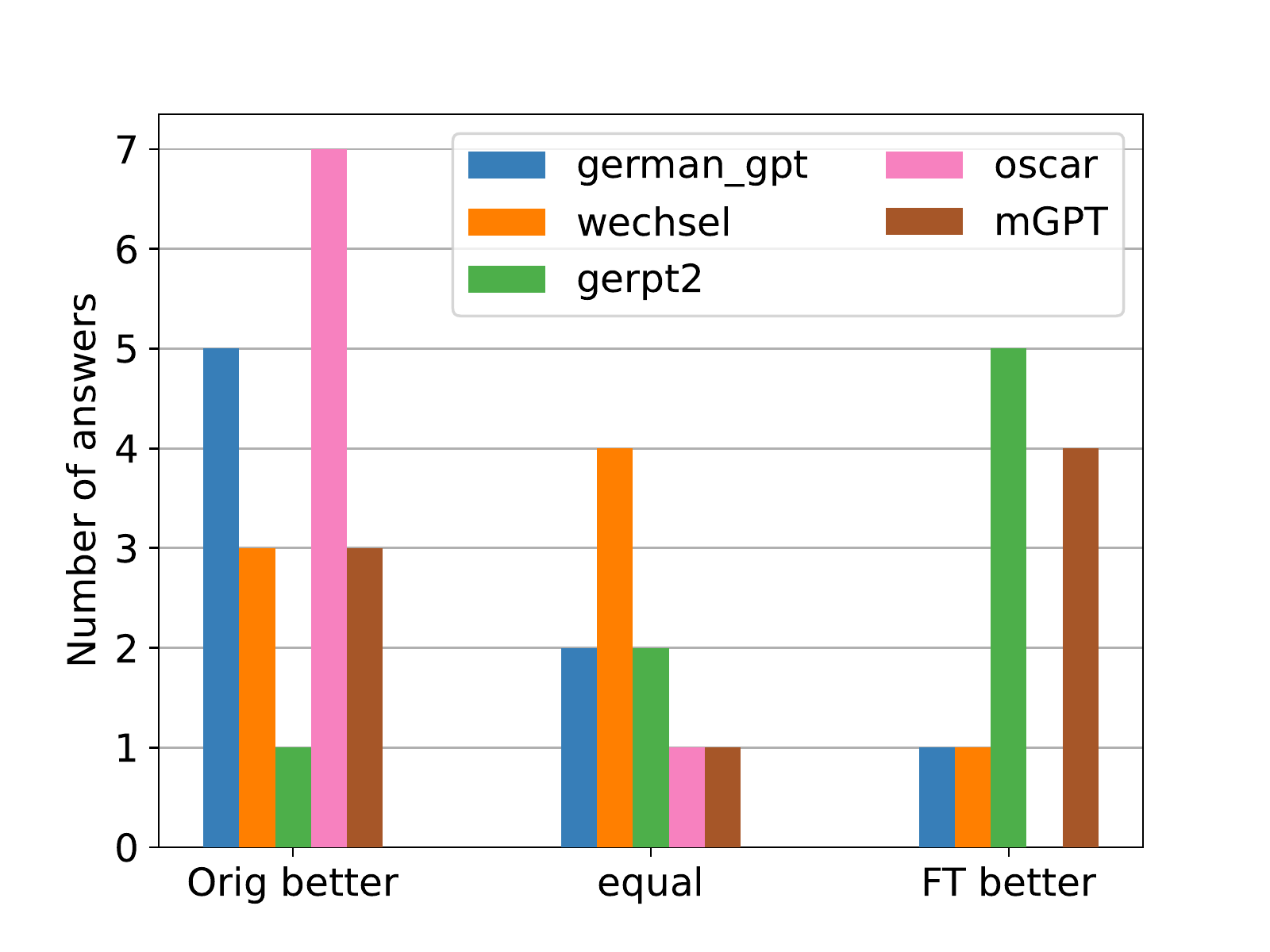}
    \caption{Human grammar evaluation with a ranking task. Participants selected which model output of the fine-tuned and original versions showed fewer grammatical mistakes.}
    \label{fig:human_grammar}
\end{figure}
Fine-tuning language models to a specific style can result in catastrophic forgetting \citep{Goodfellow-catastrophic-forgetting}. To test if our fine-tuning for Leichte Sprache influences the output quality of the models, we asked human reviewers to rate the models' grammaticality. The reviewers were not paid for their review but participated voluntarily. We selected the outputs of the prompt \enquote{Dieses Haus}(\textit{This house}) with decoding strategy contrastive from Section \ref{sec:promots}. Then, we presented the output of each original and its respective fine-tuned model side by side and asked the participants to select the candidate with fewer grammatical errors. Participants could also state that both models were equal. Overall, seven native speakers and one non-native speaker participated in the survey. The distribution of answers is shown in Figure \ref{fig:human_grammar}. While most participants preferred the fine-tuned version of gerpt2 and mGPT, the fine-tuning of oscar decreased its grammar score. When averaging over all responses and models, the worsening of the grammaticality by fine-tuning the models on Leichte Sprache is neglectable.

\subsection{Text complexity prediction}\label{subsec:text-complexity}
Fine-tuning models for a specific domain improves their performance on different tasks within this domain \citep{Gururangan-pretraining-benefits}. To test if this applies to our models, we evaluated them on the downstream task of text complexity prediction. Therefore, we added a linear layer on top of the language model heads and fine-tuned the models for the respective task. The data for this task came from the GermEval 2022 shared task on text complexity assessment \cite{Mohtaj-GermEval22}. This shared task's goal was to predict a sentence's complexity on a continuous scale between 1 and 7. We split the shared task's training data into train, evaluation, and test subsets with a ratio of 80:10:10 and fine-tuned our models for ten steps with a batch size of eight, i.e., on 80 samples total. Table \ref{tab:downstream_comparison} reports the mean squared errors on the unseen test set after the few-shot fine-tuning. The first two models have a high error for both the fine-tuned and original models. As the model only performed ten training steps, the results highly depend on the initialization. For the other three models, however, the fine-tuned models clearly outperform the original models. This gives evidence that with the fine-tuning on Easy Language data, the models get a better understanding of text complexity and, thus, can better discriminate easy from normal texts.
\begin{table}[ht]
    \centering 
    \rowcolors{2}{gray!10}{white}
    \begin{tabular}{lccc} \toprule
         & \multicolumn{3}{l}{\textbf{Mean squared error}} \\
        \multirow{-2}{*}{\textbf{Model}} && \textbf{FT} & \textbf{O}\\ \midrule
         gerpt2 && \textbf{2.36} & 4.17 \\
         german\_gpt && 6.22 & \textbf{4.25}\\
         wechsel && \textbf{0.81} & 1.79\\
         oscar && \textbf{0.83} & 1.65 \\
         mGPT && \textbf{0.92} & 1.11\\
         \bottomrule
    \end{tabular}
    \caption{Mean squared error after fine-tuning for continuous text complexity prediction on 80 sentences. Most of the fine-tuned models outperform their originals.}
    \label{tab:downstream_comparison}
\end{table}
\subsection{Text simplification}
%
We used our pre-trained language models as plug-in decoders in a mBART simplification model. As the decoders already know how to output Easy Language, we only trained the encoder-decoder cross attention. Due to computational limitations, we could not test all our language models on the text simplification downstream task. Therefore, we selected the two most promising ones, gerpt2 and german\_gpt. Table \ref{tab:simplification_comparison} shows how our simplification models perform on the 20 Minuten test dataset compared to the baseline by \citet{rios2021newdata}. To generate the simplifications, we used a beam size of four and calculated the metrics with Huggingface evaluate. Our models outperform the baseline on the SARI metric; however, they fall behind when comparing ROUGE-L and BLEU scores. All of these metrics assess how well the proposed output overlaps with a reference simplification and do not consider synonyms. SARI is a score explicitly tailored to the task of simplification, while BLEU and ROUGE-L are general translation/seq2seq metrics. Herefore, a better SARI score may be an indication that our models do more rephrasing than the baseline model and, thus, yield better simplifications. To achieve this result, our models needed training on only 7\% of the trainable parameters of the baseline while preserving state-of-the-art performance.
\begin{table}[ht]
    \centering
    \rowcolors{2}{gray!10}{white}
    \begin{tabular}{p{45pt}ccc}
    \toprule
        & & \textbf{gerpt2}& \textbf{german\_gpt}\\
        \multirow{-2}{*}{\textbf{Score}}& \multirow{-2}{*}{\textbf{Baseline*}} &\textbf{FT}&\textbf{FT}\\
        \midrule
        \mbox{\textbf{ROUGE-L}}&\textbf{19.96}&18.52&17.93\\
        \textbf{SARI}&33.29&42.25&\textbf{42.74}\\
        \textbf{BLEU}&\textbf{6.29}&4.95&4.80\\ \midrule
        \rowcolor{white}
        \textbf{\#Params}& & &\\
        \rowcolor{white}
        \textbf{trained}&\multirow{-2}{*}{\normalfont{ 416M}}& \multirow{-2}{*}{\textbf{29M}} & \multirow{-2}{*}{\textbf{29M}}\\
    \bottomrule
    \end{tabular}
    \caption{Text simplification performance on the 20 Minuten testset. For our models, only the cross attention was trained which reduced the number of trained parameters by far;\newline
    *: copied from the baseline paper \citep{rios2021newdata}.}
    \label{tab:simplification_comparison}
\end{table}
\section{Conclusion}
With this paper, we have published a collection of causal language models for German Easy Language. These models mimic the style of Easy Language and favor short and precise sentences. In addition, they adapt to the conventions of only conveying one thought per sentence and putting a line break after every sentence. We exploited these pre-trained models in a sequence-to-sequence text simplification task. As the models were already fine-tuned to the desired output style, we only had to train the encoder-decoder cross attention and, thus, reduced the number of trainable parameters by 93\%. With this, training a style-transfer system becomes feasible for settings with few aligned data or a lack of computational power. 


\section*{Limitations}
This paper focuses on the style transfer of Easy Language for German. Due to their word inflections and high average word length, languages like German are harder to learn for language models \citep{Mielke-German-hard-to-model}. Therefore, the proposed approach may work even better on easier-to-model languages, but we did not test any other language. In addition, the style transfer of simplified language uses the same vocabulary as the original language and only reduces its diversity. Our approach has yet to be evaluated on other styles, for example, ones that introduce new words.

When evaluating the influence of fine-tunung on the grammaticality of the model outputs, we found that even the original models were not perfect and produced grammatical errors. One possible reason is relying on GPT2-based models that are relatively small and, thus, perform worse than state-of-the-art language models like PaLM \citep{Chowdhery-PaLM}. In addition, the German base models are often already fine-tuned versions of English models, and thus, may already suffer from catastrophic forgetting due to fine-tuning.

\section*{Ethics Statement}
ATS systems can provide more accessible versions of texts, however, a good text simplification is targeted to the knowledge and language level of its audience. Therefore, to utilize these systems for the target group directly, the systems need to be deployed in a controllable setting where the user can set the level of simplification or ask for additional explanations if necessary. Nevertheless, there are also applications where ATS systems can increase the amount of accessible information on the internt withput being used by the target group directly. For example, these systems can yield a draft simplification for professional translators or can be helpful for public state authorities that are forced by law to offer online information in Easy Language.
Another problem is the possible stigmatization of users if they request a simplified version of the data \citep{Hansen-stigmatisation}. Finally, the availability of information in Easy Language is very sparse; thus, it is hard to fact-check material on the internet with other sources. This makes the target group of Easy Language highly vulnerable to misinformation and fake news. Hence, our generative models must be used with care as they do not provide hallucination control.

Among the sources of our dataset, there is a significant bias towards news articles as well as some regional bias due to the large proportion of articles related to Austria, Switzerland, and northern Germany. As all sources are from official website articles, and the dataset does not include user comments, we expect the data to be unoffensive and of high quality. Nevertheless, we find topical biases such as the COVID-19 pandemic due to the years from which the articles were scraped. In respect of any intellectual property laws, we published the scrapers used to obtain the data but not the data itself.

\bibliography{custom}
\bibliographystyle{acl_natbib}

\onecolumn
\appendix
\section{Overview of available data for Easy Language}
\label{sec:appendix}

\newcounter{numFootnotes}
\setcounter{numFootnotes}{3} 

\newcounter{hurrakiId}
\setcounter{hurrakiId}{\thenumFootnotes{} + 1}
\newcounter{lebenshilfeId}
\setcounter{lebenshilfeId}{\thenumFootnotes{} + 2}
\newcounter{einfachstarsId}
\setcounter{einfachstarsId}{\thenumFootnotes{} + 3}
\newcounter{nachrichtenId}
\setcounter{nachrichtenId}{\thenumFootnotes{} + 4}
\newcounter{kurierId}
\setcounter{kurierId}{\thenumFootnotes{} + 5}
\newcounter{ndrId}
\setcounter{ndrId}{\thenumFootnotes{} + 6}
\newcounter{infoeasyId}
\setcounter{infoeasyId}{\thenumFootnotes{} + 7}

\begin{table*}[h!]
\centering
    \rowcolors{2}{gray!10}{white}
\begin{tabular}{p{3,5cm} p{1cm} p{1,25cm} p{8cm}}
\toprule
\textbf{Dataset} & \textbf{Articles} & \textbf{Sentences} & \textbf{Description} \\
\midrule
 Hurraki\protect\footnotemark[\thehurrakiId{}]              & 3,911 & 56,785      & Wikipedia-style dictionary\\
 Lebenshilfe\protect\footnotemark[\thelebenshilfeId{}]          & 396   & 7,144       & Dictionary for people with intellectual disabilities  \\
 Einfachstars\protect\footnotemark[\theeinfachstarsId{}]         & 6,488 & 129,674     & News about celebrities    \\
 Nachrichtenleicht\protect\footnotemark[\thenachrichtenId{}]   & 7,709 & 122,842     & News published by Deutschlandfunk \\
 Kurier\protect\footnotemark[\thekurierId{}]              & 4,519 & 67,827      & News for Austria  \\
 NDR\protect\footnotemark[\thendrId{}]                & 1,817 & 60,749      & News for the states of Lower Saxony, Mecklenburg-Vorpommern, and Schleswig-Holstein \\
 InfoEasy\protect\footnotemark[\theinfoeasyId{}]              & 163 & 10,310      & News for Switzerland \\
 \citet{trondheim:2018} & 44  & 4,210       & Compilation of election programs, excerpts from the Bible, children's stories, and Red Cross documents   \\ 
 \bottomrule
\end{tabular}
\caption{Overview of the available monolingual data in Easy Language.}
\label{tab:monolingualDataOverview}
\end{table*}

\footnotetext[\thehurrakiId{}]{\url{https://hurraki.de/}}
\footnotetext[\thelebenshilfeId{}]{\url{https://www.lebenshilfe.de/woerterbuch}}
\footnotetext[\theeinfachstarsId{}]{\url{https://einfachstars.info/}}
\footnotetext[\thenachrichtenId{}]{\url{https://www.nachrichtenleicht.de/}}
\footnotetext[\thekurierId{}]{\url{https://kurier.at/einfache-sprache}}
\footnotetext[\thendrId{}]{\url{https://www.ndr.de/fernsehen/barrierefreie_angebote/leichte_sprache}}
\footnotetext[\theinfoeasyId{}]{\url{https://infoeasy-news.ch/}}

\vspace{-.3cm}

\newcounter{brandeinsId}
\setcounter{brandeinsId}{\thenumFootnotes{} + 8}
\newcounter{gruenenId}
\setcounter{gruenenId}{\thenumFootnotes{} + 9}
\newcounter{mdrnewsId}
\setcounter{mdrnewsId}{\thenumFootnotes{} + 10}
\newcounter{mdrdictId}
\setcounter{mdrdictId}{\thenumFootnotes{} + 11}

\begin{table*}[h!]
\centering
\rowcolors{6}{gray!10}{white}
\begin{tabularx}{\linewidth}{p{3cm}ccX}
\toprule
\textbf{Dataset} & \textbf{Articles} & \textbf{Sentences} & \textbf{Description}\\
\midrule
 Kurier\protect\footnotemark[\thekurierId{}]           & 3,476    & -      & Article-aligned news data from Austria\\
 BrandEins\protect\footnotemark[\thebrandeinsId{}]        & 212       & -      & Paragraph-aligned data from a business journal\\
 Wahlprogramm: Die Gr\"{u}nen\protect\footnotemark[\thegruenenId{}]       & -          &  100         & Sentence-wise manually-aligned data from the election program of the Green party\\
 MDR news\protect\footnotemark[\themdrnewsId{}]          & -          & 100 & Sentence-wise manually-aligned data from the news for the states of Thuringia, Saxony, and Saxony-Anhalt\\
 MDR dictionary\protect\footnotemark[\themdrdictId]     &  -         & 100         & Manually-aligned data of dictionary entries between MDR Easy Language entries and German Wikipedia articles\\
 \citet{rios2021newdata} &  18,305    & -       & Full articles paired with simplified summaries from the Swiss news magazine 20 Minuten\\
 \citet{Sauberli-first-NMT} &  -    & 19,724    & Sentence-aligned news data from Austria Press Agency aligned using CATS \cite{vstajner2018cats}\\
 \citet{Toborek2022Sep} &  708 & 5,942      & Both article and sentence-aligned compilation of texts on health and medication, public administration, politics, information texts for disabled people, and news articles (has some overlap with some sources listed in Table~\ref{tab:monolingualDataOverview})    \\
 \citet{aumiller2022klexikon}   & 2,898 & -      & German online encyclopedia for children, called Klexikon (it contains simplified concepts rather than Easy Language)\\
 \bottomrule
\end{tabularx}
\caption{Overview of the parallel data in simplified German and Easy Language.}
\label{tab:parallelDataOverview}
\end{table*}

\footnotetext[\thebrandeinsId{}]{\url{https://www.brandeins.de/themen/rubriken/leichte-sprache}}
\footnotetext[\thegruenenId{}]{\url{https://www.gruene-bw.de/wahlen/landtagswahl-2021/wahlprogramm/wahlprogramm-in-leichter-sprache/}}
\footnotetext[\themdrnewsId{}]{\url{https://www.mdr.de/nachrichten-leicht/index.html}}
\footnotetext[\themdrdictId{}]{\url{https://www.mdr.de/nachrichten-leicht/woerterbuch/index.html}}

\end{document}